\documentclass{article}
\usepackage{spconf,amsmath,graphicx}

\usepackage{graphicx}
\usepackage{amstext}
\usepackage{mathrsfs}
\usepackage{amssymb}
\usepackage{amsmath}
\usepackage{bm}
\allowdisplaybreaks[4]
\usepackage{epstopdf}
\usepackage{multicol}
\usepackage{stfloats}
\usepackage{url}
\usepackage{color}
\usepackage{enumerate}
\usepackage{breqn}
\usepackage{bbm}
\usepackage{cite}
\usepackage{caption}
\usepackage{enumitem}
\usepackage{array}
\usepackage[caption = false]{subfig}
\usepackage{multirow}
\usepackage{booktabs}
\usepackage{bookmark}
\usepackage[ruled,vlined]{algorithm2e}
\usepackage{booktabs}
\newcommand{\ie}{\textit{i.e., }}
\newcommand{\eg}{\textit{e.g., }}

\title{AI-GAN: Attack-Inspired Generation of Adversarial Examples}
\name{Tao Bai$^1$,
Jun Zhao$^1$,
Jinlin Zhu$^1$,
Shoudong Han$^2$,
Jiefeng Chen$^{3}$,
Bo Li$^{4}$,
Alex Kot$^1$}
\address{$^1$Nanyang Technological University,
$^2$Huazhong University of Science and Technology\\
$^3$University of Wisconsin-Madison,
$^4$University of Illinois at Urbana-Champaign\\}
  
\begin{document}
%
\maketitle
\begin{abstract}
Deep neural networks (DNNs) are vulnerable to adversarial examples, which are crafted by adding imperceptible perturbations to inputs.
Recently different attacks and strategies have been proposed, but how to generate adversarial examples perceptually realistic and more efficiently remains unsolved.
This paper proposes a novel framework called Attack-Inspired GAN (AI-GAN), where a generator, a discriminator, and an attacker are trained jointly.
Once trained, it can generate adversarial perturbations efficiently given input images and target classes.
Through extensive experiments on several popular datasets \eg MNIST and CIFAR-10, AI-GAN achieves high attack success rates and reduces generation time significantly in various settings.
Moreover, for the first time, AI-GAN successfully scales to complicated datasets \eg CIFAR-100 with around $90\%$ success rates among all classes.
\end{abstract}
\section{Introduction}
Deep neural networks have achieved great success in the last few years and have drawn tremendous attention from academia and industry.
With the rapid development and deployment of deep neural networks, safety concerns from society raise gradually.
Recent studies have found that deep neural networks are vulnerable to adversarial examples~\cite{DBLP:journals/corr/SzegedyZSBEGF13,DBLP:journals/corr/GoodfellowSS14}, usually crafted by adding carefully-designed imperceptible perturbations on legitimate samples.
So in human's eyes, the appearances of adversarial examples are the same as their legitimate copies, while the predictions from deep learning models are different.

Many researchers have managed to evaluate the robustness of deep neural networks in different ways, such as box-constrained \mbox{L-BFGS}~\cite{DBLP:journals/corr/SzegedyZSBEGF13}, Fast Gradient Sign Method (FGSM)~\cite{DBLP:journals/corr/GoodfellowSS14}, Jacobian-based Saliency Map Attack (JSMA)~\cite{papernot2016limitations}, C\&W attack~\cite{carlini2017towards} and Projected Gradient Descent~(PGD) attack~\cite{DBLP:conf/iclr/MadryMSTV18}.
These attack methods are optimization-based with proper distance metrics $L_{0}$, $L_{2}$ and $L_{\infty}$ to restrict the magnitudes of  perturbations and make the presented adversarial examples visually natural.
These methods are usually time-consuming, computation-intensive, and need to access the target models at the inference period for strong attacks. 

Some researchers employ generative models \eg GAN~\cite{goodfellow2014generative} to produce adversarial perturbations~\cite{XiaoLZHLS18,poursaeed2018generative,jandial2019advgan++}, or generate adversarial examples directly~\cite{NIPS2018_8052}.
Compared to optimization-based methods, generative models significantly  reduce the time of adversarial examples generation.
Yet, existing methods have two apparent disadvantages: 
1) The generation ability is limited \ie they can only perform one specific targeted attack at a time. 
Re-training is needed for different targets.
2) they can hardly scale to real world datasets. 
Most GAN based prior works evaluated their methods only on MNIST and CIFAR-10, which is not feasible for complicated reality tasks.

To solve the problems as mentioned above, we propose a new variant of GAN to generate adversarial perturbations conditionally and efficiently, which is named \mbox{Attack-Inspired} GAN (\mbox{AI-GAN}) and shown in Fig.~\ref{fig:overall_arc}. 
In AI-GAN, a generator is trained to perform targeted attacks with clean images and targeted classes as inputs; a discriminator with an auxiliary classifier for classification generated samples in addition to discrimination.
Unlike existing works, we add an attacker and train the discriminator adversarially,
which not only enhances the generator's attack ability,
but also stabilize the GAN training process~\cite{zhou2018dont,Liu_2019_CVPR}.
On evaluation, we mainly select three datasets with different classes and image sizes and compare our approach with four representative methods in white-box settings and under defences.
From the experiments, we conclude that 1) our model and loss function are useful, with much-improved efficiency and scalability;
2) our approach generates commensurate or even stronger attacks (for the most time) than existing methods under the same $L_{\infty}$ bound of perturbations in various experimental settings. 

We summarize our contributions as follows: 
\begin{enumerate}
\item Unlike prior works, We propose a novel GAN framework called \mbox{AI-GAN} in which a generator, a discriminator, and an attacker are trained jointly.
\item To our best knowledge, AI-GAN is the first GAN-based method, which generates perceptually realistic adversarial examples with different targets and scales to complicated datasets \eg CIFAR-100.
\item AI-GAN shows strong attack abilities on different datasets and outperforms existing methods in various settings through extensive experiments.
\end{enumerate}

\section{Related Work}\label{sec:Related Work} 
Adversarial examples, which are able to mislead deep neural networks, are first discovered by~\cite{DBLP:journals/corr/SzegedyZSBEGF13}.
Since then, various attack methods have been proposed.  
Authors of \cite{DBLP:journals/corr/GoodfellowSS14} developed Fast Gradient Sign Method (FGSM) to compute the perturbations efficiently using back-propagation.
One intuitive extension of FGSM is Basic Iterative Method~\cite{DBLP:journals/corr/KurakinGB16}
which executes FGSM many times with smaller $\epsilon$.
Jacobian-based Saliency Map Attack (JSMA) with $L_{0}$ distance was proposed by~\cite{papernot2016limitations}.
The saliency map discloses the likelihood of fooling the target network when modifying 
pixels in original images.
Optimization-based methods have been proposed to generate quasi-imperceptible 
perturbations with constraints in different distance metrics.
A set of attack methods are designed in~\cite{carlini2017towards}.
The objective function is minimizing $\|\delta\|_{p}+c \cdot f(x+\delta)$,
where $c$ is an constant and $p$ could be 0, 2 or $\infty$.
In \cite{Chen2017}, an algorithm was  proposed to approximate the gradients
of targeted models based on Zeroth Order Optimization (ZOO).
A convex optimization method called Projected Gradient Descent (PGD) was introduced by~\cite{DBLP:conf/iclr/MadryMSTV18} to generate adversarial examples, which is proved to be the strongest first-order attack.
However, such methods are usually time-consuming, and need to access the target model for generation.

There is another line of research working on generating adversarial examples with generative models.
Generative models are usually used to create new data  because of their powerful representation ability.
Motivated by this, Poursaeed\textit{~et~al.~}\cite{poursaeed2018generative} firstly applied generative models to generate four types of adversarial perturbations (universal or image dependent, targeted or non-targeted) with U-Net~\cite{ronneberger2015u} and ResNet~\cite{he2016deep} architectures. 
Mao\textit{~et~al.~}\cite{mao2020gap++} extended the idea of~\cite{poursaeed2018generative} with conditional targets.
Xiao\textit{~et~al.~}\cite{XiaoLZHLS18} used the idea of GAN to make adversarial examples perceptually realistic.
Different from the above methods generating adversarial perturbations, some other methods generate adversarial examples directly, which are called unrestricted adversarial examples~\cite{NIPS2018_8052}.
Song\textit{~et~al.~}\cite{NIPS2018_8052} proposed to search the latent space of pretrained ACGAN~\cite{odena2017conditional} to find adversarial examples.
Note that all these methods are only evaluated on simple datasets \eg MNIST and CIFAR-10.

\section{Our Approach}\label{sec:approach}
In this section, we first define the problem, then elaborate our proposed framework and derive the objective functions.
\subsection{Problem Definition}
Consider a classification network $f$ trained on dataset $\mathcal{X} \subseteq \mathcal{R}^n$, with $n$ being the dimension of inputs.
And suppose $(x_i, y_i)$ is the $i^{th}$ instance in the training data, where $x_i \in \mathcal{X}$ is generated from some unknown distribution $\mathcal{P}_{\mathrm{data}}$, and $y_i \in \mathcal{Y}$ is the ground truth label.
The classifier $f$ is trained on natural images and achieves high accuracy.
The goal of an adversary is to generate an adversarial example $x_{adv}$, which can fool $f$ to output a wrong prediction and looks similar to $x$ in terms of some distance metrics.
We use $L_{\infty}$ to bound the magnitude of perturbations.
There are two types of such attacks: given an instance $(x, y)$, the adversary makes $f(x_{adv}) \neq y$, which is called untargeted attack; or $f(x_{adv}) = t$ given a target class $t$, which is called targeted attack.
Targeted attacks are more challenging than untargeted attack, and we mainly focus on targeted attacks in this paper.

\subsection{Proposed Framework}
\begin{figure}[htbp]
\centering
  \includegraphics[width=\columnwidth]{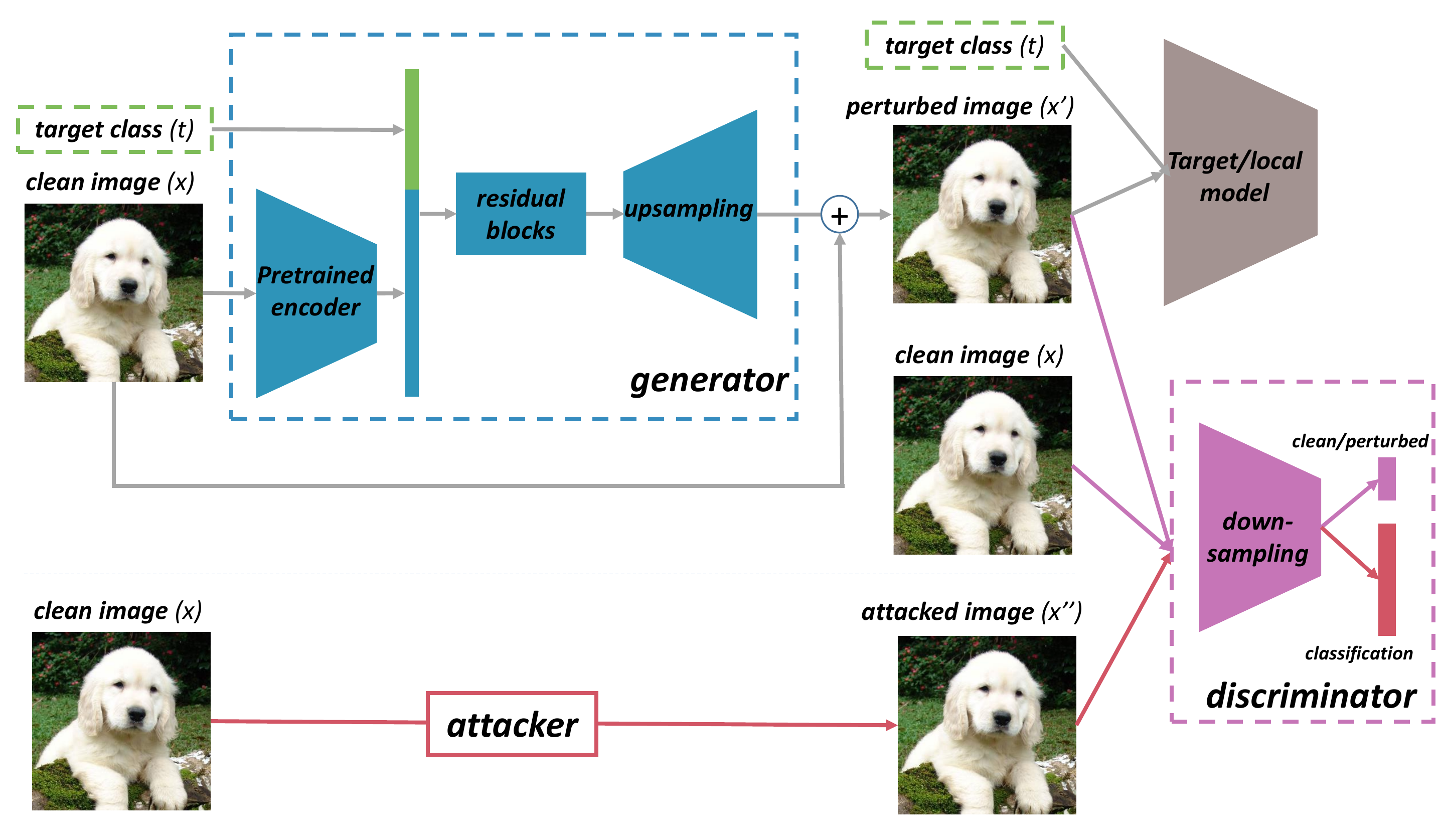}
\caption{The architecture of our AI-GAN. We use the dashed line and dot-dashed line to indicate the data flow for Stage~1 and Stage~2 respectively. Solid lines are used in both stages.}
\label{fig:overall_arc}
\end{figure}

We propose a new variant of conditional GAN called Attack-Inspired GAN (\mbox{AI-GAN}) to generate adversarial examples conditionally and efficiently.
As shown in Fig.~\ref{fig:overall_arc}, the overall architecture of \mbox{AI-GAN} consists of a generator $G$, a two-head discriminator $D$, an attacker $A$, and a target classifier $C$.
Within our approach, both generator and discriminator are trained in an end-to-end way:
the generator generates and feeds fake images to the discriminator;
Meanwhile, real images sampled from training data and their attacked copies are provided to the discriminator.
Specifically, the generator $G$ takes a clean image $x$ and the target class label $t$ as inputs to generate adversarial perturbations $G(x, t)$.
$t$ is sampled randomly from the dataset classes.
An adversarial example $x_{pert}:=x + G(x, t)$ can be obtained and sent to the discriminator $D$.
Other than the adversarial examples generated by $G$, the discriminator $D$ also takes the clean images and adversarial examples $x_{adv}$ generated by the attacker $A$.
So $D$ not only discriminates real/fake images but also classifies adversarial examples correctly.

\textbf{Discriminator.}
Different from existing methods, the discriminator of AI-GAN has two branches: one is trained to discriminate between real images $X_{real}$ and perturbed images $X_{pert}$, and another is to classify $X_{pert}$ correctly.
To further enhance the attack ability of the generator, we propose to train the classification module adversarially. 
Thus, we add an attacker into the training process.
Another benefit of a robust discriminator is that it helps stabilize and accelerate the whole training~\cite{zhou2018dont}.

Overall, the loss function of our discriminator consists of three parts: $\mathcal{L}_S$ for discriminating real/perturbed images and $\mathcal{L}_{C(adv)}$ for classification on adversarial examples generated by the attacker and the generator, which are expressed as
\begin{equation}
\begin{array}{l}
\mathcal{L}_{S}=E\left[\log P\left(S=\textit {real} \mid X_{\textit {real}}\right)\right]+ \\
\qquad \qquad \qquad E\left[\log P\left(S=\textit {pert} \mid X_{\textit {pert}}\right)\right],
\end{array}
\end{equation}
\begin{equation}
\begin{array}{l}
\mathcal{L}_{C(adv)}=E\left[\log P\left(C=y \mid X_{\textit {adv}}\right)\right],
\end{array}
\end{equation}
and
\begin{equation}
\begin{array}{l}
\mathcal{L}_{C(pert)}=E\left[\log P\left(C=y \mid X_{\textit {pert}}\right)\right],
\end{array}
\end{equation}
where $y$ represents the true label.
The goal of the discriminator is to maximize $\mathcal{L}_{S}+\mathcal{L}_{C(adv)}+\mathcal{L}_{C(pert)}$.

\textbf{Generator.}
To promote the generator's scalability, we propose to pre-train the encoder in a self-supervised way.
The pre-trained encoder can extract features effectively and reduces the training difficulties from training scratch.
A pre-trained encoder's existence makes our approach similar to feature space attacks and increases the adversarial examples' transferability somehow.
As we train a discriminator with an robust auxiliary classifier, our generator's attack ability is further enhanced.

The loss function of the generator consists of three parts: $\mathcal{L}_{target(adv)}$ for attacking target models, $\mathcal{L}_{D(adv)}$ for attacking the discriminator, and $\mathcal{L}_{S}$ same as that for the discriminator.
$\mathcal{L}_{target(adv)}$ and $\mathcal{L}_{D(adv)}$ are expressed as
\begin{equation}
\begin{array}{l}
\mathcal{L}_{target(pert)}=E\left[\log P\left(C=t \mid X_{\textit {pert}}\right)\right],
\end{array}
\end{equation}
and 
\begin{equation}
\begin{array}{l}
\mathcal{L}_{D(pert)}=E\left[\log P\left(C=t \mid X_{\textit {pert}}\right)\right],
\end{array}
\end{equation}
where $t$ is the class of targeted attacks.
The goal of the generator is to maximize $\mathcal{L}_{target(pert)}+\mathcal{L}_{D(pert)}-\mathcal{L}_{S}$.

\section{Experimental Results}\label{sec:experiment}
In this section, we conduct extensive experiments to evaluate \mbox{AI-GAN}.
\textbf{First}, we evaluate the attack ability of \mbox{AI-GAN} on MNIST and CIFAR-10 in white-box settings.
\textbf{Second}, we compare AI-GAN with different attack methods with defended target models.
\textbf{Third}, we show the scalability of \mbox{AI-GAN} with on CIFAR-100.
The magnitudes of adversarial perturbations are restricted under the $L_{\infty}$ of 0.3 on MNIST and 8/255 on CIFAR-10 and CIFAR-100.
Targeted models on MNIST are model A from~\cite{DBLP:conf/iclr/TramerKPGBM18} and model B from~\cite{carlini2017towards};
For \mbox{CIFAR-10}, we use ResNet32 and Wide ResNet34 (short for WRN34)~\cite{he2016deep,zagoruyko2016wide}.
In general, AI-GAN shows impressive performances on attacks and improves the computation efficiency as demonstrated in Table~\ref{tab:com_4_attacks}.
\begin{table}[!htb]
\caption{Comparison with the state-of-the-art attack methods.}
\label{tab:com_4_attacks}
\resizebox{\columnwidth}{!}{%
\begin{tabular}{cccccc}
\toprule
& FGSM  & C\&W             & PGD                      & AdvGAN           & AI-GAN           \\
\midrule
Run Time         & 0.06s & \textgreater{}3h & \multicolumn{1}{c}{0.7s} & \textless{}0.01s & \textless{}0.01s \\
\bottomrule
\end{tabular}%
}
\end{table}

\subsection{White-box Attack Evaluation}
Attacking in white-box settings is the worst case for target models where the adversary knows everything about the models.
This subsection evaluates AI-GAN on MNIST and \mbox{CIFAR-10} with different target models.
The attack success rates of \mbox{AI-GAN} are summarized in Table~\ref{table: semi-whitebox w/o d}.

\begin{figure}[htb]
  \subfloat[Adversarial examples and their perturbations on MNIST.]{%
  \begin{minipage}{\columnwidth}
  \includegraphics[width=.24\textwidth]{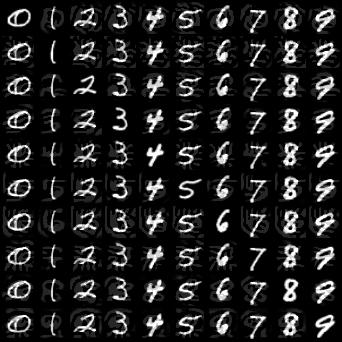}\hfill
  \includegraphics[width=.24\textwidth]{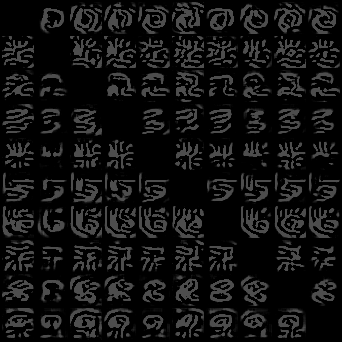}\hfill
  \includegraphics[width=.24\textwidth]{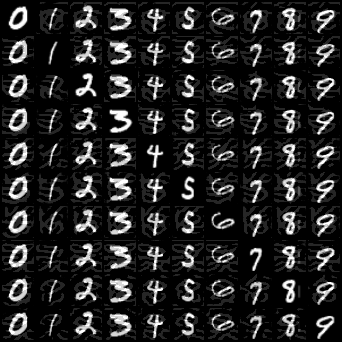}\hfill
  \includegraphics[width=.24\textwidth]{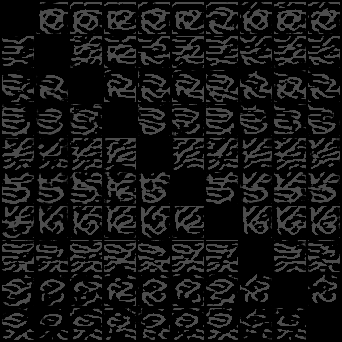}%
  \end{minipage}%
  \label{fig:wb mnist}
  }\par
  \subfloat[Adversarial examples and their perturbations on CIFAR-10.]{%
  \begin{minipage}{\columnwidth}
  \includegraphics[width=.24\textwidth]{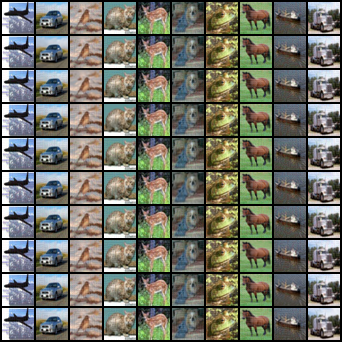}\hfill
  \includegraphics[width=.24\textwidth]{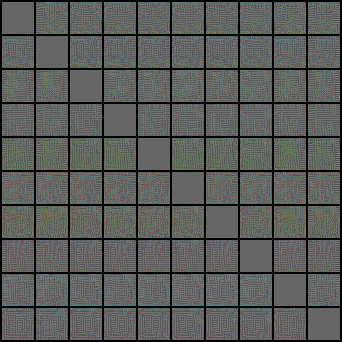}\hfill
  \includegraphics[width=.24\textwidth]{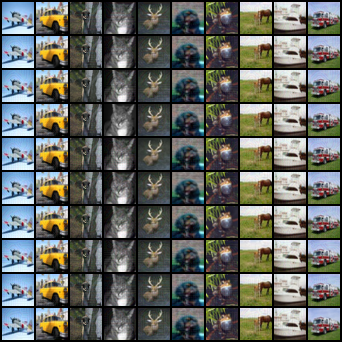}\hfill
  \includegraphics[width=.24\textwidth]{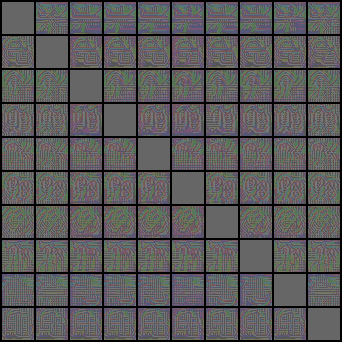}%
  \end{minipage}%
  \label{fig:wb cifar}
  }
  \caption{Visualization of Adversarial examples and perturbations generated by AI-GAN. Rows represent the different targeted classes and columns are 10 images from different classes. Original images are shown on the diagonal. Perturbations are amplified for visualization.}
   \label{fig:wb}
\end{figure}

From the table, we can see that \mbox{AI-GAN} achieves high attack success rates with different target classes on both MNIST and CIFAR-10.
On MNIST, the success rate exceeds 96\% given any targeted class.
The average attack success rates are 99.14\% for Model A and 98.50\% for Model B.
\mbox{AI-GAN} also achieves high attack success rates on \mbox{CIFAR-10}.
The average attack success rates are 95.39\% and 95.84\% for ResNet32 and 
WRN34 respectively. 
We mainly compared \mbox{AI-GAN} with AdvGAN, which is a method similar to ours.
As shown in Table~\ref{table: comparison AI-GAN and AdvGAN}, \mbox{AI-GAN} performs better than AdvGAN in most cases.
It is worth noting that \mbox{AI-GAN} can launch different targeted attacks at once, which is superior to AdvGAN.
Randomly selected adversarial examples generated are shown in Fig.~\ref{fig:wb}.
\begin{table}[htbp]
\centering
\caption{Attack success rates of adversarial examples generated by AI-GAN against 
different different models on MNIST and \mbox{CIFAR-10} in white-box settings.}
\resizebox{\columnwidth}{!}{%
\begin{tabular}{ccccc}
\toprule
  & \multicolumn{2}{c}{MNIST} & \multicolumn{2}{c}{\mbox{CIFAR-10}} \\
  \cmidrule(r){2-3} \cmidrule(r){4-5}
  Target Class & Model A   &   Model B &  ResNet32 & WRN34 \\
  \midrule
Class 1      & 98.71\%     & 99.45\%     & 95.90\%        & 90.70\%     \\
Class 2      & 97.04\%     & 98.53\%     & 95.20\%        & 88.91\%     \\
Class 3      & 99.94\%     & 98.14\%     & 95.86\%        & 93.20\%     \\
Class 4      & 99.96\%     & 96.26\%     & 95.63\%        & 98.20\%     \\
Class 5      & 99.47\%     & 99.14\%     & 94.34\%        & 96.56\%     \\
Class 6      & 99.80\%     & 99.35\%     & 95.90\%        & 95.86\%     \\
Class 7      & 97.41\%     & 99.34\%     & 95.20\%        & 98.44\%     \\
Class 8      & 99.85\%     & 98.62\%     & 95.31\%        & 98.83\%     \\
Class 9      & 99.38\%     & 98.50\%     & 95.74\%        & 98.91\%     \\
Class 10     & 99.83\%     & 97.67\%     & 94.88\%        & 98.75\%     \\
\midrule
Average      & 99.14\%     & 98.50\%     & 95.39\%        & 95.84\%     \\
\bottomrule
\end{tabular}}
\label{table: semi-whitebox w/o d}
\end{table}  

\begin{table}[htbp]
\caption{Comparison of attack success rate of adversarial examples generated by 
AI-GAN and AdvGAN in white-box setting.}
\label{table: comparison AI-GAN and AdvGAN}
\centering
\resizebox{\columnwidth}{!}{%
\begin{tabular}{ccccc}
\toprule
& \multicolumn{2}{c}{MNIST} & \multicolumn{2}{c}{\mbox{CIFAR-10}} \\
\cmidrule(r){2-3} \cmidrule(r){4-5} 
Methods & Model A   &   Model B &  ResNet32 & WRN34 \\
\midrule
AdvGAN & 97.90\% & 98.30\% & \textbf{99.30\%} & 94.70\% \\
AI-GAN & \textbf{99.14\%} & \textbf{98.50\%} & 95.39\% & \textbf{95.84\%} \\
\bottomrule
\end{tabular}}
\end{table}

\subsection{Attack Evaluation Under defenses}
In this subsection, we evaluate our method in the scenario, where the victims are aware of the potential attacks.
So defenses are employed when training targeted models.
There are various defenses proposed against adversarial examples in the literature~\cite{DBLP:conf/iclr/MadryMSTV18,samangouei2018defensegan},
and adversarial training~\cite{DBLP:conf/iclr/MadryMSTV18} is widely accepted as the most effective way~\cite{pang2020bag}.
From these defense methods, we select three popular adversarial training methods to improve the robustness of target models: (1) Adversarial training with FGSM~\cite{DBLP:journals/corr/GoodfellowSS14}, (2) Ensemble Adversarial Training~ \cite{DBLP:conf/iclr/TramerKPGBM18}, and (3) Adversarial Training with PGD~\cite{DBLP:conf/iclr/MadryMSTV18}.
However, the adversaries don't know the defenses and will use the vanilla target models in white-box settings as their targets.

We compared \mbox{AI-GAN} with FGSM, C\&W attack, PGD attack, and AdvGAN quantitatively against these defense methods, and the results are summarized in \mbox{Table~\ref{table: semi-whitebox w/ d}}. 
As we can see, \mbox{AI-GAN} has the highest attack success rates and nearly outperforms all other approaches.
\begin{table}[htbp]
\caption{Comparison of attack success rates of adversarial examples generated by 
different methods in whitebox setting with defenses. The top-2 success rates are in bold text.}
\label{table: semi-whitebox w/ d}
\resizebox{\columnwidth}{!}{%
\begin{tabular}{cccccccc}
\toprule
Dataset                  & Model                      & Defense  & FGSM    & C\&W             & PGD              & AdvGAN           & AIGAN            \\
\midrule
\multirow{6}{*}{MNIST}   & \multirow{3}{*}{Model A}   & Adv.     & 4.30\%  & 4.60\%           & \textbf{20.59\%} & 8.00\%           & \textbf{23.85\%} \\
                         &                            & Ens.     & 1.60\%  & 4.20\%           & \textbf{11.45\%} & 6.30\%           & \textbf{12.17\%} \\
                         &                            & Iter.Adv & 4.40\%  & 3.00\%           & \textbf{11.08\%} & 5.60\%           & \textbf{10.90\%} \\\cline{2-8}
                         & \multirow{3}{*}{Model B}   & Adv.     & 2.70\%  & 3.00\%           & 10.67\%          & \textbf{18.70\%} & \textbf{20.94\%} \\
                         &                            & Ens.     & 1.60\%  & 2.20\%           & 10.34\%          & \textbf{13.50\%} & \textbf{10.73\%} \\
                         &                            & Iter.Adv & 1.60\%  & 1.90\%           & 9.90\%           & \textbf{12.60\%} & \textbf{13.12\%} \\
\midrule
\multirow{6}{*}{CIFAR10} & \multirow{3}{*}{Resnet 32} & Adv.     & 5.76\%  & 8.35\%           & 9.22\%           & \textbf{10.19\%} & \textbf{9.85\%}  \\
                         &                            & Ens.     & 10.09\% & 9.79\%           & \textbf{10.06\%} & 8.96\%           & \textbf{12.48\%} \\
                         &                            & Iter.Adv & 1.98\%  & 0.02\%           & \textbf{11.41\%} & 9.30\%           & \textbf{9.57\%}  \\\cline{2-8}
                         & \multirow{3}{*}{WRN 34}    & Adv.     & 0.10\%  & 8.74\%           & 8.09\%           & \textbf{9.86\%}  & \textbf{10.17\%} \\
                         &                            & Ens.     & 3.00\%  & \textbf{12.93\%} & 9.92\%           & 9.07\%           & \textbf{11.32\%} \\
                         &                            & Iter.Adv & 1.00\%  & 0.00\%           & \textbf{9.87\%}  & 8.99\%           & \textbf{9.91\%} 
                         \\
\bottomrule
\end{tabular}%
}
\end{table}

\subsection{Scalability of AI-GAN}
\begin{figure}[htbp]
\centering
  \includegraphics[width=0.8\columnwidth]{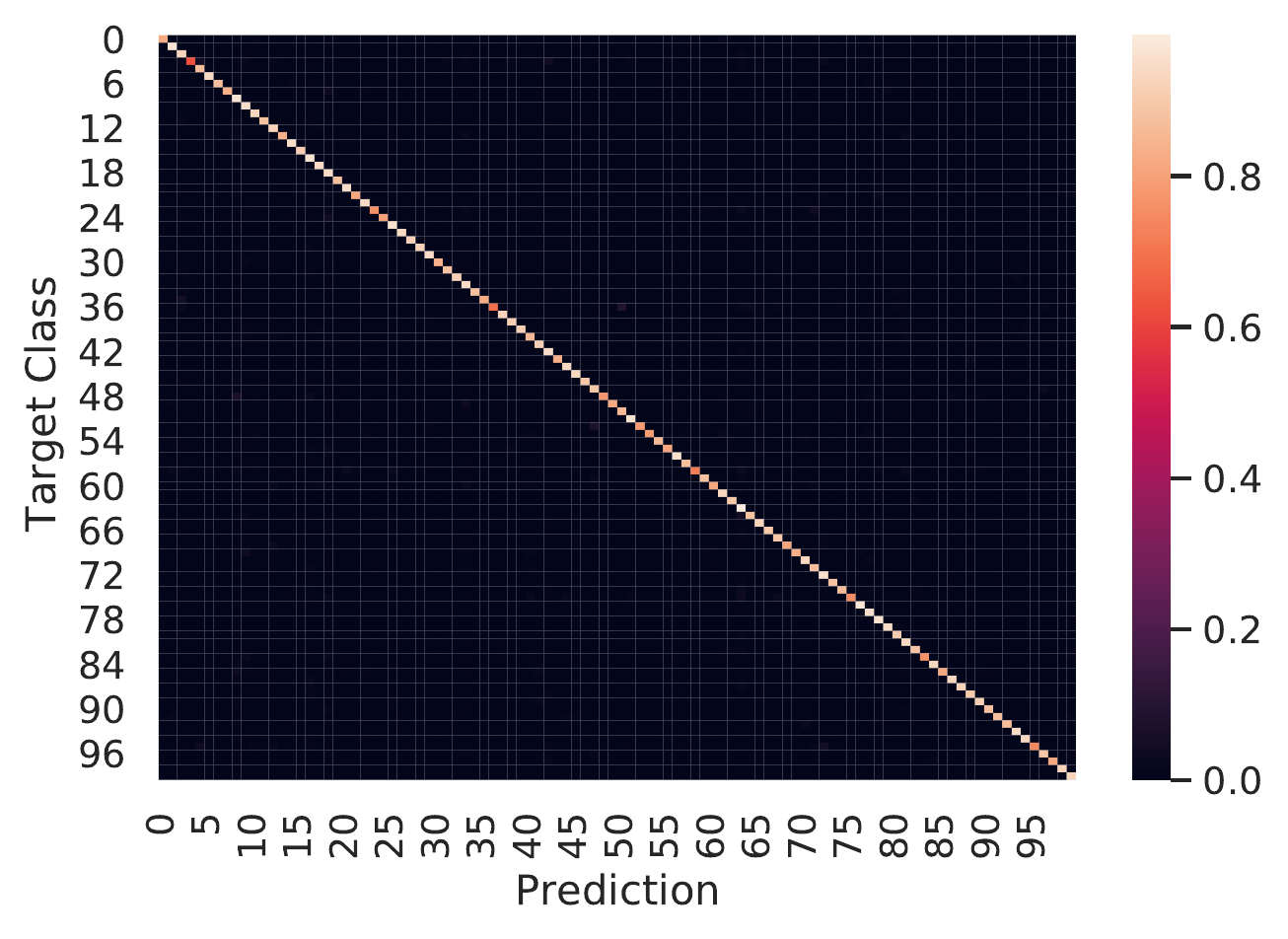}
  \caption{Visualization of attack success rates of AI-GAN on CIFAR-100 with different target classes. As it is targeted attack, the diagonal line looks lighter.}
  \label{fig: attack large dataset} 
\end{figure}

One concern of our approach is whether it can generalize to complicated datasets?
This section demonstrates the effectiveness and scalability of our approach to CIFAR-100, which is more complicated than CIFAR-10 and MNIST.
All the attacks in our experiments are targeted, so we visualized the confusion matrix in Fig.~\ref{fig: attack large dataset}, where the rows are target classes, and columns are prediction.
The diagonal line in Fig.~\ref{fig: attack large dataset} shows the attack success rate for each targeted class.
We can see all the attack success rates are very high, and the average for all classes is 87.76\%.

\section{Conclusion}\label{sec:con}
In this paper, we propose \mbox{AI-GAN} to generate adversarial examples conditionally, where we train a generator, a discriminator, and an attacker jointly.
Once AI-GAN is trained, it can perform adversarial attacks with different targets, which significantly promotes efficiency and preserves image quality.
We compare AI-GAN with several SOTA methods under different settings \eg white-box or defended, and AI-GAN shows comparable or superior performances.
With the novel architecture and training objectives, AI-GAN scales to large datasets successfully.

\bibliographystyle{IEEEbib}
\bibliography{bibliography}

\begin{thebibliography}{10}

\bibitem{DBLP:journals/corr/SzegedyZSBEGF13}
Christian Szegedy, Wojciech Zaremba, Ilya Sutskever, Joan Bruna, Dumitru Erhan,
  Ian~J Goodfellow, and Rob Fergus,
\newblock ``{Intriguing Properties of Neural Networks},''
\newblock in {\em ICLR}, 2014.

\bibitem{DBLP:journals/corr/GoodfellowSS14}
Ian~J Goodfellow, Jonathon Shlens, and Christian Szegedy,
\newblock ``{Explaining and Harnessing Adversarial Examples},''
\newblock in {\em ICLR}, 2015.

\bibitem{papernot2016limitations}
Nicolas Papernot, Patrick McDaniel, Somesh Jha, Matt Fredrikson, Z~Berkay
  Celik, and Ananthram Swami,
\newblock ``{The Limitations of Deep Learning in Adversarial Settings},''
\newblock in {\em EuroS{\&}P}. IEEE, 2016, pp. 372--387.

\bibitem{carlini2017towards}
Nicholas Carlini and David Wagner,
\newblock ``{Towards Evaluating the Robustness of Neural Networks},''
\newblock in {\em SP}, 2017, pp. 39--57.

\bibitem{DBLP:conf/iclr/MadryMSTV18}
Aleksander Madry, Aleksandar Makelov, Ludwig Schmidt, Dimitris Tsipras, and
  Adrian Vladu,
\newblock ``{Towards Deep Learning Models Resistant to Adversarial Attacks},''
\newblock in {\em ICLR}, 2018.

\bibitem{goodfellow2014generative}
Ian~J Goodfellow, Jean Pouget-Abadie, Mehdi Mirza, Bing Xu, David Warde-Farley,
  Sherjil Ozair, Aaron Courville, and Yoshua Bengio,
\newblock ``{Generative Adversarial Nets},''
\newblock in {\em NIPS}, 2014, pp. 2672--2680.

\bibitem{XiaoLZHLS18}
Chaowei Xiao, Bo~Li, Jun-Yan Zhu, Warren He, Mingyan Liu, and Dawn Song,
\newblock ``{Generating Adversarial Examples with Adversarial Networks},''
\newblock in {\em IJCAI}, 2018.

\bibitem{poursaeed2018generative}
Omid Poursaeed, Isay Katsman, Bicheng Gao, and Serge Belongie,
\newblock ``Generative adversarial perturbations,''
\newblock in {\em Proceedings of the IEEE Conference on Computer Vision and
  Pattern Recognition}, 2018, pp. 4422--4431.

\bibitem{jandial2019advgan++}
Surgan Jandial, Puneet Mangla, Sakshi Varshney, and Vineeth Balasubramanian,
\newblock ``Advgan++: Harnessing latent layers for adversary generation,''
\newblock in {\em ICCV Workshops}, 2019.

\bibitem{NIPS2018_8052}
Yang Song, Rui Shu, Nate Kushman, and Stefano Ermon,
\newblock ``{Constructing Unrestricted Adversarial Examples with Generative
  Models},''
\newblock in {\em NIPS}, pp. 8312--8323. 2018.

\bibitem{zhou2018dont}
Brady Zhou and Philipp Krähenbühl,
\newblock ``Don't let your discriminator be fooled,''
\newblock in {\em International Conference on Learning Representations}, 2019.

\bibitem{Liu_2019_CVPR}
Xuanqing Liu and Cho-Jui Hsieh,
\newblock ``{Rob-GAN: Generator, Discriminator, and Adversarial Attacker},''
\newblock in {\em CVPR}, 2019.

\bibitem{DBLP:journals/corr/KurakinGB16}
Alexey Kurakin, Ian~J Goodfellow, and Samy Bengio,
\newblock ``{Adversarial Examples in the Physical World},''
\newblock {\em CoRR}, 2016.

\bibitem{Chen2017}
Pin-Yu Chen, Huan Zhang, Yash Sharma, Jinfeng Yi, and Cho-Jui Hsieh,
\newblock ``{ZOO: Zeroth Order Optimization based Black-box Attacks to Deep
  Neural Networks without Training Substitute Models},''
\newblock in {\em AISec}, 2017, pp. 15--26.

\bibitem{ronneberger2015u}
Olaf Ronneberger, Philipp Fischer, and Thomas Brox,
\newblock ``{U-net: Convolutional Networks for Biomedical Image
  Segmentation},''
\newblock in {\em MICCAI}. Springer, 2015, pp. 234--241.

\bibitem{he2016deep}
Kaiming He, Xiangyu Zhang, Shaoqing Ren, and Jian Sun,
\newblock ``{Deep Residual Learning for Image Recognition},''
\newblock in {\em CVPR}, 2016, pp. 770--778.

\bibitem{mao2020gap++}
Xiaofeng Mao, Yuefeng Chen, Yuhong Li, Yuan He, and Hui Xue,
\newblock ``Gap++: Learning to generate target-conditioned adversarial
  examples,''
\newblock {\em arXiv preprint arXiv:2006.05097}, 2020.

\bibitem{odena2017conditional}
Augustus Odena, Christopher Olah, and Jonathon Shlens,
\newblock ``{Conditional Image Synthesis with Auxiliary Classifier GANs},''
\newblock in {\em ICML}. JMLR. org, 2017, pp. 2642--2651.

\bibitem{DBLP:conf/iclr/TramerKPGBM18}
Florian Tram{\`{e}}r, Alexey Kurakin, Nicolas Papernot, Ian~J Goodfellow, Dan
  Boneh, and Patrick~D McDaniel,
\newblock ``{Ensemble Adversarial Training: Attacks and Defenses},''
\newblock in {\em ICLR}, 2018.

\bibitem{zagoruyko2016wide}
Sergey Zagoruyko and Nikos Komodakis,
\newblock ``Wide residual networks,''
\newblock {\em arXiv preprint arXiv:1605.07146}, 2016.

\bibitem{samangouei2018defensegan}
Pouya Samangouei, Maya Kabkab, and Rama Chellappa,
\newblock ``Defense-gan: Protecting classifiers against adversarial attacks
  using generative models,''
\newblock in {\em International Conference on Learning Representations}, 2018.

\bibitem{pang2020bag}
Tianyu Pang, Xiao Yang, Yinpeng Dong, Hang Su, and Jun Zhu,
\newblock ``Bag of tricks for adversarial training,'' 2020.

\end{thebibliography}

\end{document}